\documentclass{article}
\usepackage{iclr2027_conference,times}
\usepackage{amsmath,amsfonts,bm}

\def\eqref#1{equation~\ref{#1}}
\def\1{\bm{1}}

\DeclareMathAlphabet{\mathsfit}{\encodingdefault}{\sfdefault}{m}{sl}
\SetMathAlphabet{\mathsfit}{bold}{\encodingdefault}{\sfdefault}{bx}{n}

\usepackage{graphicx}
\usepackage{booktabs}
\usepackage{enumitem}
\usepackage{hyperref}
\usepackage{url}

\title{\raisebox{-0.12em}{\includegraphics[height=1.05em]{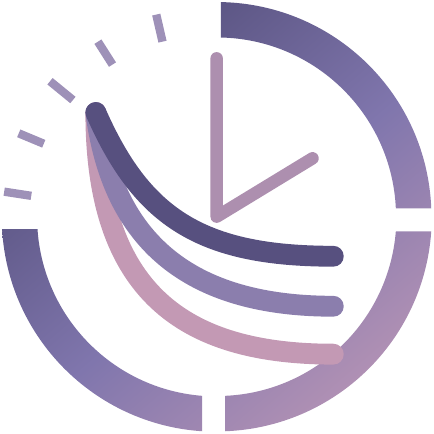}}\hspace{0.35em}A Unified Scaling Law for\\Time Series Foundation Models}
\author{%
Xilin Dai$^{1,2}$, Yiding Liu$^{1}$, Zewei Dong$^{1,*}$, Jiang-Ming Yang$^{1}$, Qiang Xu$^{2}$\\
\textnormal{$^{1}$Ant International}\\
\textnormal{$^{2}$The Chinese University of Hong Kong}\\
\textnormal{\texttt{\{xldai26,qxu\}@cse.cuhk.edu.hk}}\\
\textnormal{\texttt{zewei.dong@ant-intl.com}}\\
\textnormal{$^{*}$Corresponding author}
}
\hypersetup{pdftitle={A Unified Scaling Law for Time Series Foundation Models},pdfauthor={Xilin Dai, Yiding Liu, Zewei Dong, Jiang-Ming Yang, Qiang Xu}}

\iclrfinalcopy
\begin{document}
\maketitle
\begin{abstract}
We develop a Unified Scaling Law and a Unified Theory of Time Series Learning to understand how model capacity and historical information support forecasting. Across different lookback lengths and forecast horizons, we analyze 18,768 experimental cells from 21 checkpoints on 23 dataset-frequency tasks spanning six domains. Our empirical methodology integrates local resource relations into a parsimonious, fitted five-parameter law: capacity gains increase with history, context gains diminish toward saturation, and horizon effects enter as a common shift. Fitted without Toto~2.0, the law predicts its horizon-averaged capacity-scaling curves with mean absolute percentage errors of 1.09\% and 1.50\% at input lengths 2048 and 4096. To understand how history supports prediction, our learning theory uses Gaussian regression to analyze rule identification and predictive capability. We hypothesize that full-shot models learn by \emph{accumulating information in weights}, while frozen time series foundation models (TSFMs) use history by \emph{extracting information through activations}. Matched-history comparisons establish the predictive value of additional history. Controlled parameter exchanges and activation interventions provide evidence that history-derived rule information can be retained, reused across queries, and used to recover a contribution to long-context prediction. Together, these findings inform capacity scaling, context allocation, and the development of models that retain and apply historical rules.
Code and main results are available at \url{https://github.com/Fifthky/UniScale}.
\end{abstract}
\section{Introduction}
\label{sec:introduction}

Time series foundation models (TSFMs), large neural models pretrained from scratch on time series, are becoming more capable as forecasting research expands \citep{chronos2024,timesfm2024,moirai2024,toto2026,liu2026falcon}. Public benchmarks such as GIFT-Eval and fev-bench bring together a growing range of forecasting models \citep{gift2024,fev2025}, while application studies investigate their use in financial forecasting, precipitation prediction, and battery degradation \citep{financefoundation2026,precipitation2026,battery2026}. This progress has intensified debate over the value and scalability of foundation forecasting. Recent critiques question the premise of domain universality \citep{categoryerror2026}. Empirical studies show that selected supervised baselines can compete with TSFMs under particular tasks and history budgets \citep{specialized2025,parroting2026}. Alongside these critiques, early scaling studies establish benefits from increasing training resources and identify interactions between model complexity and history \citep{shi2024scaling,yao2025towards,edwards2024scaling}. This raises the question: \textbf{under what resource and forecasting conditions does pretrained scale translate into useful predictive gains?} Existing work establishes training-resource scaling \citep{yao2025towards,edwards2024scaling}, characterizes interactions of capacity and lookback \citep{shi2024scaling}, and studies how longer contexts support forecasting \citep{timerxl2025,longcontext2026}. However, these works have not fully resolved the ongoing debate because their experiments remain limited in either scale (model and dataset diversity) or scope (pretrained capacity, available history, and forecasting horizon). This debate calls for more general and scalable quantification across heterogeneous TSFMs to analyze error under a wide range of tasks.

Therefore, we propose a \textbf{Unified Scaling Law}, aiming to deepen understanding of how TSFMs turn pretrained capacity and history information into useful forecasts, and to inform their future training and application. We fitted the Law through controlled experiments. The protocol yields 18,768 dataset--configuration--checkpoint cells and 1,070 aggregated response points from 21 checkpoints on 23 dataset-frequency tasks across six domains. Public results from GIFT-Eval provide cross-source evaluation on the same task panel. Our empirical methodology uses local resource relations and Occam's razor to guide law construction, parameter estimation, and predictive validation. As a case study, we withhold the Toto 2.0 series of five same-architecture checkpoints, spanning approximately 4M to 2.5B parameters, and use the law estimated from the remaining heterogeneous checkpoints to predict its capacity-scaling response. The Mean Absolute Percentage Errors (MAPEs) of the horizon-averaged capacity curves are 1.09\% and 1.50\% with input lengths of 2048 and 4096, showing the Law remains informative when applied to a held-out architecture.

Further, to understand why and how TSFMs work and how they relate to full-shot models, we propose a \textbf{Unified Theory of Time Series Learning}. History carries information about both the current state and the rule governing its evolution. We hypothesize that models trained on the target series (full-shot models) learn by \emph{accumulating information in weights}, while frozen TSFMs use history by \emph{extracting information through activations}, inspired by earlier work \citep{russin2025parallel,zhang2025zeroshotchaotic,specialized2025}. We analyze this learning problem through Gaussian regression and investigate the hypotheses through experiments matching the full-shot training-record length to the TSFM input length, complemented by controlled tests of rule identification, parameter retention, and cross-query activation transfer. The empirical methodology and learning theory provide a reusable basis for future scaling research as models and forecasting conditions evolve.

Our main contributions are:
\begin{itemize}[noitemsep, topsep=0pt, leftmargin=*]
\item \textbf{A unified scaling law and empirical research methodology.} 
We develop a Unified Scaling Law linking capacity, input length, and output horizon through extensive experiments. We contribute an empirical methodology from local resource analysis to law construction, estimation, and validation on external public results and a held-out case study.
\item \textbf{A unified theory of time series learning.} We combine Gaussian regression analysis with hypotheses about how TSFMs and full-shot models turn historical evidence into predictive rules. Matched-history comparisons and controlled parameter and activation interventions examine how rule information is retained and applied to new queries.
\item \textbf{Guidance for model builders and users.} We use the Unified Scaling Law and learning analysis to give model developers empirical support for pretraining larger models. We guide users in selecting models, input lengths, and adaptation strategies for their forecasting requirements and deployment constraints.
\end{itemize}

\section{Related work}
\label{sec:related}

\noindent\textbf{Towards Time Series Foundation Models.}
Forecasting has progressed from statistical methods to Transformer and linear models \citep{patchtst2023,dlinear2023,dai2026samples}, and to TSFMs pretrained across datasets \citep{chronos2024,timesfm2024,moirai2024,toto2026}. Benchmarks compare these models \citep{gift2024,fev2025,xu2026fideltshighfidelitymultimodalbenchmark,chen2026numericaltimeseriesunified}, alongside debate over domain universality \citep{categoryerror2026}. Comparisons with supervised models and history-based predictors highlight the roles of available history and adaptation \citep{specialized2025,zhang2025zeroshotchaotic,ready2025,parroting2026,dai2026learningcontexterrorsblackbox}. Studies of in-weight learning and in-context inference offer a perspective on information use \citep{russin2025parallel,faw2025incontext}. We connect learning hypotheses with matched-history experiments and the Unified Scaling Law to study how TSFMs use forecasting resources. Our cross-query interventions examine whether historical rule information remains useful on independent queries.

\noindent\textbf{Scaling laws for language and time series models.}
Language-model scaling laws relate loss to parameters, data, and computation \citep{kaplan2020scaling,hoffmann2022training}, with test-time scaling extending resource allocation to inference \citep{snell2025testtime}. Time series studies examine training-resource scaling and capacity--lookback relations \citep{shi2024scaling,yao2025towards,edwards2024scaling}, while others study predictability and history requirements \citep{longcontext2026,wang2026exploringaccuracylawdeep,huang2026goodlinearmodelstimeseries}. Timer-XL, Toto~2.0, and YingLong investigate longer contexts, capacity-dependent horizon stability, and output scaling, respectively \citep{timerxl2025,toto2026,yinglong2025}. These studies establish resource effects in different settings but leave their joint description across heterogeneous released checkpoints unresolved. Our Unified Scaling Law connects pretrained capacity, input length, and the final scored horizon. Detailed related works are provided in Appendix~\ref{app:detailed_related}.

\section{Problem definition and evidence design}
\label{sec:problem}

We select a fixed panel $\mathcal D$ of 23 dataset-frequency pairs from GIFT-Eval based on available history and coverage across six domains, spanning 15 base datasets. Eligible evaluation cells provide more than 1,000 pre-origin observations. Appendix~\ref{app:protocol} lists the tasks and coverage.

Mean absolute scaled error (MASE) \citep{hyndman2006accuracy} is a principal point-forecasting metric in GIFT-Eval \citep{gift2024}. Let $m$ identify a checkpoint, $d\in\mathcal D$ a dataset-frequency pair, $N$ the checkpoint's active parameter count in millions, $L$ the available input length supplied to the model, and $H$ the number of scored future steps. We define the normalized response $R$ as raw MASE divided by that of Seasonal Naive (SN) at matching forecast origins and horizons:
\begin{equation}
R_{m,d,H,L}=
\frac{\operatorname{MASE}^{\mathrm{raw}}_{m,d,H,L}}
{\operatorname{MASE}^{\mathrm{raw}}_{\mathrm{SN},d,H}}.
\label{eq:relative}
\end{equation}
Throughout the paper, figures, and tables, \emph{MASE} denotes this normalized quantity $R$; values below one indicate lower error than SN. This dimensionless reference makes equal-weight averages interpretable across datasets with different units and magnitudes. Raw MASE uses the benchmark's seasonal scale and each forecast's median. Continuous calculations use data-supported input lengths; discrete displays group them into nominal levels, labelled simply as input length (Appendix~\ref{app:response_points}). We call $(N,H,L)$ NHL coordinates and use natural logarithms.

The Controlled Grid-Horizon Panel varies input length and horizon at shared forecast origins; the Controlled GIFT-Horizon Panel retains native benchmark horizons and origins. Grid observations take precedence at overlapping conditions. Equal dataset means per checkpoint and resource condition, merged at repeated actual NHL coordinates within that checkpoint, yield 1,070 Controlled fitting points (Table~\ref{tab:evidence}). Removing overlapping checkpoint and resource bins from GIFT Public leaves 111 test points from 22 checkpoints. Appendix~\ref{app:protocol} specifies grids, aggregation, and overlap removal.

\begin{table}[t]
\caption{Grid-first selection retains dataset-level results (cells); points aggregate responses for fitting and prediction. Controlled GIFT-Horizon contains only cells absent from Controlled Grid-Horizon.}
\label{tab:evidence}
\centering
\small
\setlength{\tabcolsep}{4pt}
\begin{tabular*}{\linewidth}{@{\extracolsep{\fill}}lrrrr@{}}
\toprule
Source & TSFM Checkpoints & Dataset-Frequency Task & Experimental Cells & Points \\
\midrule
Controlled GIFT-Horizon & 21 & 23 & 3,128 & 390 \\
Controlled Grid-Horizon & 21 & 23 & 15,640 & 680 \\
Combined Controlled & 21 & 23 & 18,768 & 1,070 \\
GIFT Public & 37 & 23 & 2,257 & 199 \\
\bottomrule
\end{tabular*}
\end{table}

\section{Deriving the Unified Scaling Law}
\label{sec:law}

Let $R(N,H,L)$ denote the aggregate response surface, and define log resources $n=\log N$, $h=\log H$, and $\ell=\log L$. The log-response surface is $f(n,h,\ell)=\log R(e^n,e^h,e^\ell)$.

\paragraph{Local resource relations.}
We summarize capacity contrasts at common horizon and context conditions across heterogeneous checkpoints. Within a checkpoint and horizon, adjacent input lengths provide context contrasts, represented by a first-order dependence of context elasticity on log length. Shared-origin horizon changes estimate horizon elasticity at fixed checkpoint and input. Let $\beta$ denote the capacity--history coupling coefficient, $\gamma$ the context coefficient, $q$ the context-curvature coefficient, and $\alpha$ the horizon elasticity. These local approximations give
\begin{align}
\frac{\partial f}{\partial n}&=-\beta\ell,\label{eq:capacity}\\
\frac{\partial f}{\partial\ell}&=-\gamma-\beta n+2q\ell,\label{eq:context}\\
\frac{\partial f}{\partial h}&=\alpha.\label{eq:horizon}
\end{align}
For $\beta>0$, capacity gains increase with log history length; $q>0$ represents diminishing context gains, with consistent mixed derivative $-\beta$. An affine extension adds a capacity-elasticity intercept $\lambda$, giving $\lambda-\beta\ell$ and an additional $\lambda n$ term after integration. Equation~\ref{eq:capacity} selects the zero-intercept form for parsimony; Section~\ref{sec:form_selection} compares both forms and flexible alternatives. The horizon elasticity $\alpha$ describes changes relative to the matching baseline, whose error also depends on $H$. Appendix~\ref{app:horizon_interactions} compares this common shift with $nh$ and $\ell h$ extensions.

\paragraph{Integration and the context saturation scale.}
Integrating these local approximations gives the five-parameter form with response intercept $c$ (Appendix~\ref{app:derivation}):
\begin{equation}
\boxed{\log R=c+\alpha\log H-\gamma\log L
-\beta\log N\log L+q(\log L)^2.}
\label{eq:law}
\end{equation}
For $q>0$, define the operational \emph{context saturation scale} $L_\star(N)$ as the input length where the fitted context elasticity reaches zero:
\begin{equation}
L_\star(N)=\exp\!\left(\frac{\gamma+\beta\log N}{2q}\right).
\label{eq:lstar}
\end{equation}
The fitted quadratic's stationary minimum in $\log L$ locates this scale: beyond this region, additional history yields little average improvement and may slightly increase error within the observed range through 8192 (Appendix~\ref{app:long_context_support}). Under the common-horizon-shift specification, $H$ changes response height while $N$ changes the saturation scale.

\paragraph{Estimation.}
We estimate the local relations before setting the response level. Capacity slopes at common horizon and context conditions are pooled across checkpoint subsets, and adjacent context contrasts hold checkpoint and horizon fixed. A robust objective fits $(\gamma,\beta,q)$ to these targets over $\mathbb R^3$. We estimate $\alpha$ as the median matched horizon elasticity in the Controlled Grid-Horizon Panel and $c$ as the median residual over the combined response points. Dataset-held-out and complete-series tests repeat every estimation step after excluding the held-out observations. Appendix Sections~\ref{app:derivation} and~\ref{app:estimation} specify the contrasts, weighting, and optimization.

\section{Systematic evidence for joint resource scaling}
\label{sec:experiments}

\subsection{Unified response and observed resource profiles}
\begin{figure}[t]
\centering
\includegraphics[width=\linewidth]{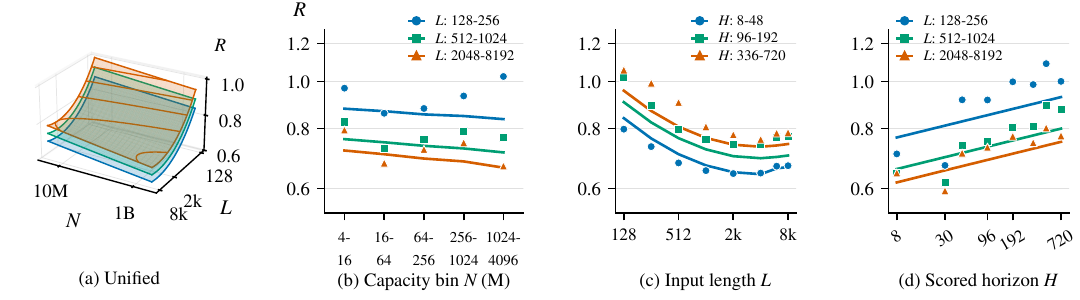}
\caption{The Unified Scaling Law and observed resource profiles. (a) Predicted MASE over $N,L$ at $H=48,192,720$ (blue, green, orange). (b)--(d) Observed means (markers) and mean predictions (lines) over identical capacity, input-length, and horizon groups, using all 1,070 Controlled points. Resource axes are logarithmic; $R$ is linear in (a) and logarithmic in (b)--(d). 1k denotes 1,024 steps.}
\label{fig:response}
\end{figure}

The combined controlled evidence gives the predicted response $\widehat R$:
\begin{equation}
\begin{split}
\log \widehat R={}&0.965780+0.042866\log H-0.362138\log L\\
&-0.001846\log N\log L+0.022217(\log L)^2.
\end{split}
\label{eq:fitted}
\end{equation}
Figure~\ref{fig:response} compares this prediction with observed capacity, context, and horizon profiles.

\subsection{Selecting a parsimonious law for the resource relations}
\label{sec:form_selection}

\begin{figure}[t]
\centering
\includegraphics[width=\linewidth]{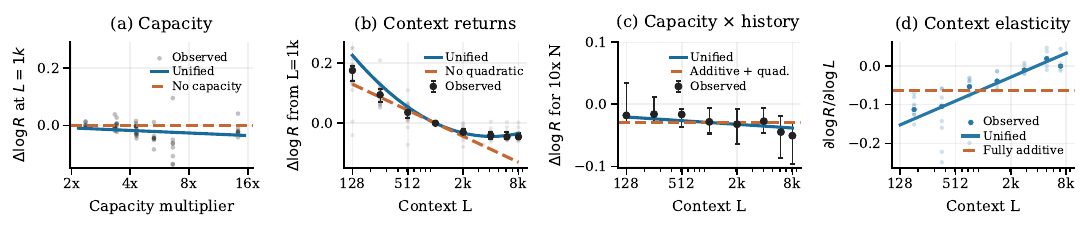}
\caption{Observed resource relations and reduced-law restrictions: (a) capacity response, (b) changing context returns, (c) history-dependent capacity gains, and (d) context elasticity for the representative 22M group. Unified curves use fitted coefficients; orange dashed curves impose restrictions without refitting. Faint points show individual Controlled contrasts; emphasized markers show medians, with interquartile-range (IQR) bars where shown. Appendix~\ref{app:form_geometry} specifies each comparison.}
\label{fig:reduced_law_properties}
\end{figure}

\paragraph{Part I: resource properties and reduced laws.}
The observed contrasts motivate three properties: capacity dependence, changing context returns, and capacity--history coupling. With subscripts on $f$ denoting partial derivatives, removing capacity or curvature forces $f_n=0$ or $f_{\ell\ell}=0$, respectively. Replacing the interaction with an additive capacity term forces $f_{n\ell}=0$, making capacity gains independent of history and the context saturation scale independent of capacity. A fully additive form imposes both restrictions. Figure~\ref{fig:reduced_law_properties} compares these restrictions with Controlled observations; Appendix~\ref{app:form_geometry} specifies contrasts; Appendix~\ref{app:balanced_context} verifies the context trend on fixed complete support.

\paragraph{Part II: quantitative comparison under a parsimony prior.}
Occam's razor guides selection among forms that retain the observed properties: when predictive advantages are small, we prefer fewer parameters and fewer auxiliary shape assumptions. The product $n\ell$ is the lowest-degree interaction in log resources, and $\ell^2$ is the lowest-degree term with nonzero context curvature, with no additional transition scale. A shallow minimum and a smooth plateau can express observed diminishing context gains through 8192. We re-estimate each candidate under the same dataset-held-out protocol to compare the Unified law with flexible alternatives and horizon-interaction extensions. Appendix Sections~\ref{app:comparison} and~\ref{app:horizon_interactions} give candidate forms, evaluation procedure, and supporting results.

\subsection{Predictive validation}
\label{sec:predictive_validation}

\begin{figure}[!b]
\centering
\includegraphics[width=\linewidth]{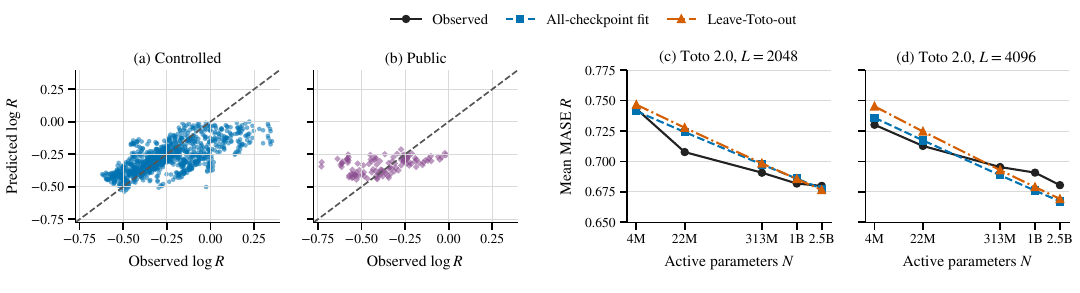}
\caption{Controlled/Public prediction and horizon-averaged Toto~2.0 capacity curves.}
\label{fig:prediction}
\end{figure}

The selected law has 10.13\% MAPE on 1,070 task-averaged Controlled response points. Applied directly to 111 non-overlapping GIFT Public points, it gives 12.82\% MAPE, testing prediction under a different source of model results and horizon--context conditions (Figure~\ref{fig:prediction}a,b). Appendix~\ref{app:prediction_stability} gives reference comparisons and estimation stability.

Toto~2.0 provides five checkpoints spanning approximately 4M--2.5B parameters \citep{toto2026}. We remove all five from both controlled sources and re-estimate the law using the remaining 16 heterogeneous checkpoints. Prediction on its 150 Controlled Grid-Horizon points gives 11.53\% MAPE. After averaging over the five horizons at each capacity, the five-point capacity curves have MAPEs of 1.09\% at $L=2048$ and 1.50\% at $L=4096$ (Figure~\ref{fig:prediction}c,d). These predictions show that a statistical relationship identified across heterogeneous model designs can capture scaling within a held-out architecture. Appendix Figure~\ref{fig:toto_full} gives the complete Toto evaluation.

\section{A Unified Theory of Time Series Learning}
\label{sec:mechanism}
\newcommand{\MechanismTaskCount}{1186}
\newcommand{\MechanismFitCount}{108}
\newcommand{\MechanismTotalFitCount}{1212}
\newcommand{\MechanismCrossedFitCount}{1104}
\newcommand{\MechanismFrozenConditionCount}{36}
\newcommand{\MechanismTransferConditionCount}{18}
\newcommand{\MechanismParameterConditionCount}{54}
\newcommand{\MechanismTransferReduction}{7.51}
\newcommand{\MechanismRecoveryPercent}{41.67}
\newcommand{\MechanismAdaptationReduction}{14.68}
\newcommand{\MechanismChronosRecoveryReduction}{6.01}
\newcommand{\MechanismChronosTransferBefore}{0.28774}
\newcommand{\MechanismChronosTransferAfter}{0.28776}

Our Unified Theory of Time Series Learning studies how full-shot models and frozen TSFMs turn historical evidence into predictive rules. Historical evidence supports rule identification, the predictor's capabilities determine how accurately it estimates and applies the rule, and the learning path determines whether rule information is retained after training or extracted within the current forecast. Building on in-weight and in-context learning \citep{russin2025parallel,sander2024incontext}, we combine a regression analysis with two learning hypotheses and experiments on rule identification, retention, and reuse across queries.

\subsection{Historical evidence and the two learning paths}

Let $x_t$ denote the observation at time $t$, and $x_{u:v}$ the observations from indices $u$ through $v$, inclusive. For query length $\ell_0$, define history $D_{L,t}=x_{t-L+1:t}$ at forecast origin $t$ and current query $B_t=x_{t-\ell_0+1:t}$. For a first-order autoregressive process, denoted AR(1), with coefficient $\phi$, the next conditional mean is $\phi x_t$. Although the current state is $x_t$, estimating $\phi$ can require many observations, separating rule identification from current-state information \citep{longcontext2026}.

\paragraph{Full-shot hypothesis: accumulating information in weights.}
We hypothesize that training can turn history information into rule-specific parameters that remain useful across subsequent queries. With training procedure $\mathcal A$ and forecasting model $g$, write the fitted parameters $\widehat\theta_L$ and forecast $\widehat y_{\mathrm{weights}}$ as
\begin{equation}
\widehat\theta_L=\mathcal A(D_{L,t}),\qquad
\widehat y_{\mathrm{weights}}=g_{\widehat\theta_L}(B_t,H).
\label{eq:weights}
\end{equation}

\paragraph{TSFM hypothesis: extracting information through activations.}
We hypothesize that a frozen TSFM can use historical evidence to form rule-specific activation states that its subsequent computation applies to the current query. Let $F$ denote the TSFM, $\Theta_N$ its fixed pretrained parameters at capacity $N$, and $\widehat y_{\mathrm{act}}$ its forecast:
\begin{equation}
\widehat y_{\mathrm{act}}=F_{\Theta_N}(D_{L,t},H),\qquad
\Theta_N\ \text{fixed}.
\label{eq:activations}
\end{equation}

For a fixed query $B$, let $p$ be the number of rule coefficients, indexed by $j=1,\ldots,p$, with fixed features $\psi_j(B)$ and coefficient prior variances $\tau_j^2$. Writing $\mathcal N(\mu,v)$ for a Gaussian distribution with mean $\mu$ and variance $v$, take independent rule coefficients $\omega_j\sim\mathcal N(0,\tau_j^2)$ and independent innovation $\varepsilon$ with variance $\sigma^2$, distributed as $\mathcal N(0,\sigma^2)$. The scalar target is $Y=\sum_{j=1}^{p}\psi_j(B)\omega_j+\varepsilon$. With per-observation evidence precision $\nu_j>0$, history supplies independent Gaussian observations of these coefficients with noise precision $L\nu_j$. A predictor uses posterior means for the first $r\le p$ available coefficients and prior means for the remainder. Gaussian conditioning \citep{rasmussen2006gaussian} gives its prediction-error variance $V_r(L)$ and expected absolute error $\mathcal E_r(L)$:
\begin{equation}
V_r(L)=\sigma^2+
\sum_{j\le r}\frac{\psi_j(B)^2\tau_j^2}{1+L\nu_j\tau_j^2}
+\sum_{j>r}\psi_j(B)^2\tau_j^2,
\qquad
\mathcal E_r(L)=\sqrt{\frac{2V_r(L)}{\pi}}.
\label{eq:learning_variance}
\end{equation}
It separates uncertainty reducible by history from predictive variation unused by this predictor.

In this model, the benefit of an additional useful coefficient increases with history length: more evidence improves its estimate. With positive remaining variance, the gains from further history diminish at sufficiently long inputs. The same rule estimate can be retained after training or computed from current history by a frozen predictor. For a single coefficient, training that leaves less uncertainty reduces the gain from additional immediate evidence about the same rule. Enlarging the available coefficient set creates opportunities to use history, while refining an existing estimate reduces the need for further evidence about it. Appendix~\ref{app:learning_theory} gives the assumptions and proofs.

\subsection{Predictive value of training history and immediate input}

\begin{figure}[t]
\centering
\includegraphics[width=\linewidth]{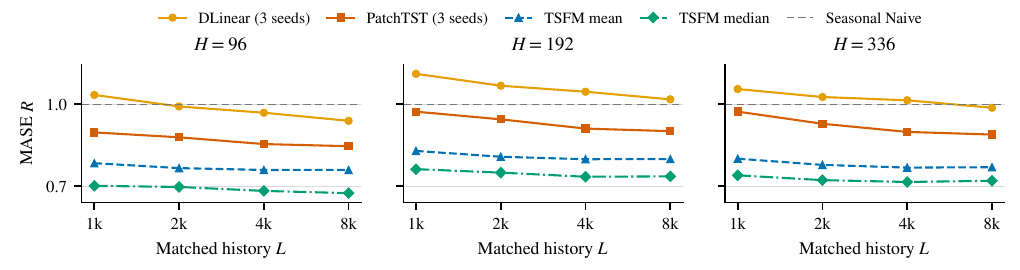}
\caption{Forecasting with matched history budgets on 23 tasks. Full-shot curves average three seeds; TSFM curves show the mean and median over a fixed set of available checkpoints covering the grid. Seed error bars are too small to distinguish at this scale: mean pointwise relative standard deviations are $\pm0.208\%$ for DLinear and $\pm0.502\%$ for PatchTST. The line marks Seasonal Naive.}
\label{fig:mechanism}
\end{figure}

We compare DLinear and PatchTST \citep{dlinear2023,patchtst2023} with frozen TSFMs on 23 tasks at $L\in\{1024,2048,4096,8192\}$ and $H\in\{96,192,336\}$. Full-shot models train on pre-origin history with local input $\ell_0=96$ and retain fitted states during scoring; TSFMs receive length-$L$ histories at the same origins. This varies the shared history budget in earlier comparisons \citep{zhang2025zeroshotchaotic,specialized2025}; Appendix~\ref{app:mechanism_history} details implementation and training. Both full-shot architectures improve task-averaged MASE at every tested history extension and horizon (Figure~\ref{fig:mechanism}). At $H=96$, increasing history from 1,024 to 8,192 reduces DLinear's MASE from 1.042 to 0.931 and PatchTST's from 0.885 to 0.833. The fixed TSFM panel's mean and median remain lower at every matched condition; PatchTST's gaps to both narrow over this history range at all three horizons.

A complementary crossed experiment varies training history $T\in\{4096,8192\}$ and immediate input $C\in\{96,512\}$ independently at $H=96$, on the same 23 tasks with three seeds. Alongside DLinear and two PatchTST sizes, we compare the same TimesFM 2.5 checkpoint before and after task-specific adaptation, with each input configuration trained and validation-selected separately. MASE falls from 0.9233 to 0.7877 under adaptation, a \MechanismAdaptationReduction{}\% reduction in the task-averaged score. Appendix~\ref{app:mechanism_crossed} reports the results. The MASE gap between the 96-step and 512-step TimesFM configurations decreases from 0.2396 before adaptation to 0.0768 afterwards, a reduction of about 68\%. This supports partial substitution between retaining information through training and supplying longer immediate input, as illustrated by the single-rule analysis in Appendix~\ref{app:learning_substitution}.

\subsection{Rule identification and retention}

We use three stationary processes: Gaussian AR(1), Gaussian lag-8, and a nonlinear threshold process. Within each process, two rule signs $s\in\{-1,+1\}$ share a single-observation distribution with mean zero and variance one. The rule changes the temporal transition relation. For each test query, the last 96 observations are identical under both rules; conditionally generated earlier histories provide evidence for the alternative transitions. Independent queries include both signs and multiple amplitudes, so applying a rule requires combining historical information with the new query.

We evaluate one-step predictions against the known conditional mean using \emph{conditional-mean MASE}, measuring rule agreement without future innovation noise. The naive reference uses the last query observation and the same target. Appendix~\ref{app:mechanism_design} specifies the processes, sampling, scoring, and independent evidence splits.

\paragraph{Historical evidence and rule identification.}
TimesFM 2.5 and Chronos 2 improve conditional-mean MASE from 96 to 2,048 observations, with positive paired-query 95\% bootstrap intervals in all \MechanismFrozenConditionCount{} model--process--magnitude--normalization conditions. Fixed population statistics preserve the gains. An oracle knowing the process family and magnitude, and inferring only the sign from the historical prefix, reaches conditional-mean MASE below 0.001 at rule magnitude $a=0.15$ and $L=512$ for all three processes (Appendix~\ref{app:mechanism_context}). With queries and marginal distributions fixed, historical transition structure permits accurate rule identification, while TSFM errors and rule responses continue to vary with input length.

\paragraph{Trained parameters retain rules for new queries.}
We train DLinear and a compact PatchTST separately under each rule using three history lengths and three seeds, for \MechanismFitCount{} fits. Each validation-selected model then predicts independent queries using only their 96 current observations. Parameters are exchanged between opposite-rule models, with normalization buffers calibrated on a shared independent history mixture. Predictions move toward the donor rule in all \MechanismParameterConditionCount{} process--architecture--length--seed conditions. Both architectures also improve their mean forecast accuracy as training history increases. At $L=8192$, both architectures attain conditional-mean MASE below 0.13 on the Gaussian processes, averaged over three seeds (Appendix~\ref{app:mechanism_weights}).

\subsection{Rule transfer and recovery through activations}

We intervene inside frozen TimesFM 2.5 using activation patching \citep{zhang2024patching}. We average four independent donors' intermediate-state differences between 512 observations and their final 96 observations. The primary layer, token span, history length, and strength are fixed across processes and experiments. For \emph{transfer}, we add this historical state difference to the short-input state of a new query. For \emph{recovery}, we replace the corresponding state in the query's natural long-context execution with its short-input state, then add the same donor difference; other long-context states remain unchanged. Transfer tests cross-query reuse; recovery tests restoration of its contribution to natural long-context prediction.

\begin{table}[t]
\caption{Shared-site transfer and recovery in TimesFM 2.5. Conditional-mean MASE, averaged over three rule magnitudes and two donor repetitions. Natural, erased, and recovered predictions share the 512-step recipient history.}
\label{tab:mechanism_transfer}
\centering
\small
\begin{tabular}{lrrrrr}
\toprule
Process & Query only & Transfer & Natural & Erased & Recovered \\
\midrule
AR(1) & 0.26289 & 0.24181 & 0.17697 & 0.23986 & 0.21229 \\
Lag-8 & 0.26972 & 0.25106 & 0.19702 & 0.25074 & 0.23755 \\
Threshold & 0.30168 & 0.27873 & 0.24983 & 0.28404 & 0.26195 \\
\bottomrule
\end{tabular}

\end{table}

Transfer reduces mean error by \MechanismTransferReduction{}\%; independent-donor recovery restores \MechanismRecoveryPercent{}\% of the mean error increase caused by erasure at the selected site. Erasure raises error, and transfer and recovery improve it with positive paired-query 95\% intervals in all \MechanismTransferConditionCount{} conditions. Both interventions outperform random and carrier-only controls; rule-dependent gains also hold against shuffled-history donors (Appendix~\ref{app:mechanism_decomposition}). With recipient query and history fixed, matching-rule donors outperform opposite-rule donors for both interventions in every tested TimesFM condition across three layers and three strengths. Across the nine layer--strength settings, mean recovery improves over erasure in all nine and mean direct transfer over query-only prediction in seven. Chronos 2 shows the same matching advantage on independent confirmation data at a development-selected setting: recovery lowers mean error by \MechanismChronosRecoveryReduction{}\%, with positive intervals in all 18 conditions, while direct transfer leaves overall error nearly unchanged. Appendix~\ref{app:mechanism_specificity} reports both models' controls and intervention robustness.

\begin{figure}[htbp]
\centering
\includegraphics[width=\linewidth]{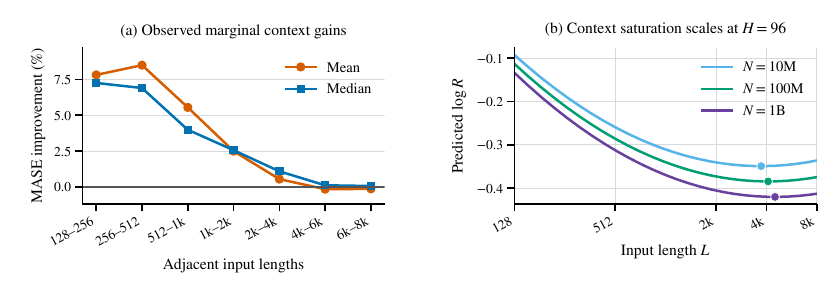}
\caption{Context gains and fitted saturation scales. Left: mean and median MASE improvements between adjacent inputs for matched checkpoint--horizon pairs. Right: fitted $\log R$ at $H=96$ for three capacities; $L_\star$ markers locate approximate saturation regions within the evaluated range.}
\label{fig:resources}
\end{figure}

\section{Practical guidance for model builders and users}
\label{sec:implications}

\subsection{For model builders: coordinate capacity, context, and horizon}

The Unified Scaling Law links capacity, input length, and forecast horizon, providing a basis for joint model and training design. For the fitted $\beta,q>0$, capacity gains at fixed horizon and input length and the capacity dependence of the saturation scale satisfy
\begin{equation}
\frac{\partial\log R}{\partial\log N}=-\beta\log L,
\qquad
\frac{d\log L_\star}{d\log N}=\frac{\beta}{2q}>0.
\label{eq:capacity_guidance}
\end{equation}
Increasing capacity lowers fitted error and shifts saturation toward longer histories, providing empirical support for pretraining larger models. Because capacity gains depend on history, compare model sizes across the input lengths intended for deployment. Training and validation should cover the application's input--horizon combinations. The horizon term describes performance relative to the matched baseline, so validation should retain horizon-specific references when assessing these gains. Alongside average error, evaluate how models use historical information. The controlled tests in Section~\ref{sec:mechanism} assess whether historical rules are identified, retained, and applied across queries. An oracle can distinguish insufficient historical evidence from inaccurate use of available evidence; measuring rule responses together with forecast error helps identify improvements in history use.

\subsection{For model users: select models and allocate history}

Start from the required forecast horizon and available history, then compare model capacity and input length jointly. The fitted capacity relation and Toto~2.0 observations support considering larger checkpoints, particularly when longer histories are available \citep{toto2026}. For fixed capacity and horizon, the law gives
\begin{equation}
\frac{\partial\log R}{\partial\log L}
=2q\log\!\frac{L}{L_\star(N)}.
\label{eq:context_guidance}
\end{equation}
Context gains weaken near the fitted saturation scale, providing a broad starting region for task-level validation (Figure~\ref{fig:resources}). Evaluate candidate models and contexts at the required horizon under the deployment latency and memory constraints; when errors are similar, shorter inputs reduce both costs. Seasonal Naive provides a reference for deciding when a TSFM offers useful gains, including conditions with short histories or long horizons. A frozen TSFM is a natural starting point when task-specific training is unavailable; full-shot training can retain information from a long record for repeated short-input forecasts. The crossed experiment motivates validating adaptation together with deployment input length. Retune context after adaptation using the adapted model's own validation curve: information learned during training can make shorter-input configurations more competitive.

\section{Discussion and conclusion}
\label{sec:conclusion}

\paragraph{Discussion.}
The Unified Scaling Law describes average resource returns, while our learning analysis distinguishes evidence available in history from its predictive use. Retention and extraction provide complementary ways to apply historical rules. Task-specific adaptation can shift the useful balance between retained information and immediate input. These findings motivate evaluating history-use capability alongside capacity and jointly validating training and deployment choices.

\paragraph{Conclusion.}
We establish a five-parameter Unified Scaling Law through observed resource properties, parsimony, and validation on Public results and held-out Toto~2.0 checkpoints. Our Unified Theory of Time Series Learning combines Gaussian regression with matched-history comparisons and controlled interventions. Parameter exchanges support retained rule information, while cross-query activation transfer and recovery support its reuse in frozen prediction. The empirical methodology provides a reusable sequence from local resource analysis to estimation and validation. Together with the learning theory, it supports future scaling studies as models and forecasting settings evolve, while informing capacity scaling, context allocation, and model development.

\newpage
\bibliography{iclr2027_conference}
\bibliographystyle{iclr2027_conference}
\clearpage
\appendix
\section{Detailed related works}
\label{app:detailed_related}

\subsection{Time series foundation models}

Time series modeling supports diverse applications, including financial forecasting, precipitation prediction, and battery monitoring \citep{financefoundation2026,precipitation2026,battery2026,dai_socnet_2025}.

Time series forecasting has expanded from statistical methods to Transformer-based models such as PatchTST and iTransformer \citep{patchtst2023,itransformer2024} and linear models such as DLinear, FITS, and TimePrism \citep{dlinear2023,fits2024,dai2026samples}, alongside the recent rise of large-scale pretraining \citep{kottapalli2025foundationmodelstimeseries}. Chronos casts probabilistic prediction as token modeling \citep{chronos2024}, TimesFM develops a decoder-only forecaster \citep{timesfm2024}, and Moirai introduces unified training across datasets and frequencies \citep{moirai2024}. Subsequent releases include Chronos-Bolt and Chronos-2 \citep{chronos2024,chronos2_2025}, Moirai-MoE and Moirai~2 \citep{moirai_moe2025,moirai2_2025}, and the TiRex and TiRex-2 forecasting models \citep{tirex2025,tirex2_2026}. Other models include Sundial \citep{sundial2025}, Granite FlowState \citep{flowstate2026}, and Granite PatchTST-FM \citep{patchtstfm2026}, alongside the Toto and Toto~2.0 series \citep{toto2025,toto2026}. Time-MoE and Timer-S1 explore expert-based scaling \citep{time_moe2025,timers1_2026}, while Timer-XL develops expanded forecasting contexts \citep{timerxl2025}; other released forecasters include Falcon-X, Kairos, YingLong, and Reverso \citep{liu2026falcon,kairos2026,yinglong2025,reverso2026}. The public comparison also covers TabPFN-TS \citep{tabpfnts2026}, and the forecasting releases t0, CleanTS, goia, PatchFM, and LiteSpecFormer \citep{t0_2026,cleants,goia,patchfm,litespecformer2026}. GIFT-Eval and fev-bench provide comparisons across heterogeneous domains, frequencies, and forecast lengths \citep{gift2024,fev2025}.

\subsection{Scaling laws for language and time series models}

Language-model scaling laws characterize loss through parameters, training data, and computation \citep{kaplan2020scaling,hoffmann2022training}. Test-time scaling extends the allocation problem to inference \citep{snell2025testtime}. For time series, \citet{shi2024scaling} jointly analyze data size, model complexity, and lookback with a fixed forecast horizon, finding that useful lookback depends on the training-data regime. \citet{yao2025towards} study parameter, data, and compute scaling across in-domain and out-of-domain forecasting, while \citet{edwards2024scaling} study training-resource laws with a fixed context. Complementary studies examine predictability and context selection. \citet{longcontext2026} explain why identifying a generating process can require history longer than its memory, while \citet{sander2024incontext} analyze how Transformers estimate autoregressive dynamics from context. \citet{wang2026exploringaccuracylawdeep} relate the lowest forecasting errors attained by deep models to window-wise pattern complexity, while \citet{huang2026goodlinearmodelstimeseries} show that the relationship between useful lookback and forecast horizon varies across datasets for tuned linear models. 

Timer-XL examines longer contexts and the dependence of their benefits on model and training choices \citep{timerxl2025}. Toto~2.0 demonstrates improved forecasting performance with increasing model size and examines how capacity affects long-horizon stability \citep{toto2026}. YingLong studies output scaling \citep{yinglong2025}. Diversified inference investigates sampling and perturbations alongside capacity and context length \citep{diversified2026}. These findings concern different architectures and inference procedures, without establishing a shared error law across heterogeneous released checkpoints. 

\subsection{Analysis and controversy surrounding TSFMs}

The growth of foundation forecasting has prompted scrutiny of its conceptual scope and evaluation \citep{categoryerror2026,leakage2025}. Empirical comparisons examine the performance of pretrained forecasting models across model scales and application domains \citep{chakraborty2026scaling,alonso2026pretrainedtimeseriesfoundationmodels}. \citet{specialized2025} and \citet{zhang2025zeroshotchaotic} compare frozen TSFMs with supervised baselines given the same history, using autoregressive models on forecasting benchmarks and N-BEATS on chaotic systems, respectively. \citet{ready2025} combine a representation learned by principal component analysis on a multi-domain corpus with a linear head trained on the available history, questioning the practical necessity of zero-shot prediction. Context parroting shows that matching and copying historical segments can rival TSFMs on dynamical-system forecasting and relates one-step error to history length \citep{parroting2026}. QuitoBench and FETS report context- and horizon-dependent model comparisons on application-traffic and energy data, respectively \citep{xue2026quitobenchhighqualityopentime,obermeier2026fets}. These studies establish the importance of history budgets and task-specific adaptation.

An information-use perspective connects these comparisons to learning mechanisms. \citet{russin2025parallel} distinguish incremental in-weight learning from in-context inference through activation dynamics. In forecasting, \citet{faw2025incontext} investigate adaptation through examples supplied in context and through weight fine-tuning, while \citet{longcontext2026} explain how training or inference supports identification of a generating process. Rank and compressibility analyses examine how TSFMs represent predictive information \citep{yu2026rank}. We develop learning hypotheses about history information and examine them through matched-history experiments with linear and Transformer full-shot models \citep{dlinear2023,patchtst2023}.

\section{Evaluation protocol and statistical estimation}
\label{app:protocol}

\subsection{Benchmark snapshot and common dataset scope}
We use fixed GIFT-Eval benchmark, dataset, and public-result snapshots documented in the released code. Table~\ref{tab:datasets} lists the 23 dataset-frequency pairs selected based on available history and domain coverage. They span 15 base datasets and six benchmark domains; separate frequencies constitute distinct forecasting tasks. Cloud and web operations include virtual-machine and operational monitoring, energy includes electricity demand, transformer measurements, and solar power, and environmental data include weather, air quality, and river flow. Transportation, retail sales, and births supply additional application domains. This fixed panel is shared by the controlled experiments, public analysis, Toto case study, and matched-history comparison, supplying 61 native GIFT configurations and supporting shared-origin evaluation through maximum horizon $H_{\max}=720$.

Controlled evaluations use the same evaluator with experiment-specific input cropping and forecast-origin construction. The public-inclusion filter follows the registered checkpoint catalogue, while the controlled experiment uses the checkpoint panel in Table~\ref{tab:models}. Both analysis sources retain cells whose available history exceeds 1,000 observations; controlled eligibility uses the shortest available history in the cell. This condition concerns the data record, and allocated contexts can be as short as 128 steps.

\begin{table}[htbp]
\caption{The fixed evaluation panel and its application domains. Sampling intervals span five, ten, and fifteen minutes, one hour, one day, and one week. Nineteen pairs have short, medium, and long GIFT configurations; the four daily/weekly pairs contribute four short configurations, for 61 in total.}
\label{tab:datasets}
\centering
\begin{tabular}{lll}
\toprule
Dataset & Sampling intervals & Domain \\
\midrule
Bitbrains fast storage & 5 min & Cloud/web operations \\
Bitbrains random & 5 min & Cloud/web operations \\
BizITObs L2C & 5 min, 1 h & Cloud/web operations \\
Electricity & 15 min, 1 h & Energy \\
ETT1 & 15 min, 1 h & Energy \\
ETT2 & 15 min, 1 h & Energy \\
Hierarchical sales & 1 day & Sales \\
Jena weather & 10 min, 1 h & Environment \\
KDD Cup 2018 & 1 h & Environment \\
LOOP Seattle & 5 min, 1 h & Transportation \\
M Dense & 1 h & Transportation \\
Saugeen & 1 day, 1 week & Environment \\
Solar & 10 min, 1 h & Energy \\
SZ Taxi & 15 min & Transportation \\
US births & 1 day & Healthcare \\
\midrule
15 base datasets & 23 dataset-frequency pairs & 6 domains \\
\bottomrule
\end{tabular}
\end{table}

\subsection{Checkpoint capacities and input coverage}
\label{app:models}
The controlled panel includes 21 checkpoints with heterogeneous architectures, pretraining choices, and inference interfaces. Table~\ref{tab:models} records their active and total capacities; active capacity defines the $N$ coordinate. Checkpoint names identify the released weights.

\begin{table}[htbp]
\caption{Controlled checkpoints and active and total parameter counts, in millions.}
\label{tab:models}
\centering
\small
\setlength{\tabcolsep}{4pt}
\begin{tabular*}{\linewidth}{@{\extracolsep{\fill}}lrr@{\hspace{1.5em}}lrr@{}}
\toprule
Checkpoint & Active $N$ & Total $N$ & Checkpoint & Active $N$ & Total $N$ \\
\midrule
Toto 2.0 4M & 4.14 & 4.14 & Moirai 1.1 large & 310.97 & 310.97 \\
Toto 2.0 22M & 21.92 & 21.92 & Moirai 2.0 small & 11.39 & 11.39 \\
Toto 2.0 313M & 312.68 & 312.68 & TimesFM 1.0 & 200.00 & 200.00 \\
Toto 2.0 1B & 1041.03 & 1041.03 & TimesFM 2.0 & 500.00 & 500.00 \\
Toto 2.0 2.5B & 2454.28 & 2454.28 & TimesFM 2.5 & 200.00 & 231.29 \\
Chronos 2 small & 27.93 & 27.93 & TiRex 1.1 GIFT & 35.00 & 35.00 \\
Chronos 2 & 119.48 & 119.48 & TiRex-2 GIFT-ZS & 38.40 & 82.50 \\
Chronos-Bolt small & 47.72 & 47.72 & Sundial base & 128.33 & 128.33 \\
Chronos-Bolt base & 205.29 & 205.29 & Granite FlowState r1.1 & 18.52 & 18.52 \\
Moirai 1.0 small & 13.83 & 13.83 & Granite PatchTST-FM r1 & 257.90 & 257.90 \\
Moirai 1.1 base & 91.36 & 91.36 &  &  &  \\
\bottomrule
\end{tabular*}
\end{table}

The active-capacity convention follows each checkpoint's inference path. TiRex-2 uses 38.4M active parameters out of 82.5M total for univariate inference \citep{tirex2_2026}. TimesFM \citep{timesfm2024} capacities follow the metadata of its 2.0 and 2.5 checkpoints, with the latter's additional quantile head included in the total count alongside its 200M backbone. Nominal capacities such as 200M and 500M preserve the precision available in the metadata. Frozen inference settings follow each checkpoint's benchmark profile; exact configurations, context limits, and snapshot provenance are provided in the released code.

\begin{figure}[htbp]
\centering
\includegraphics[width=\linewidth]{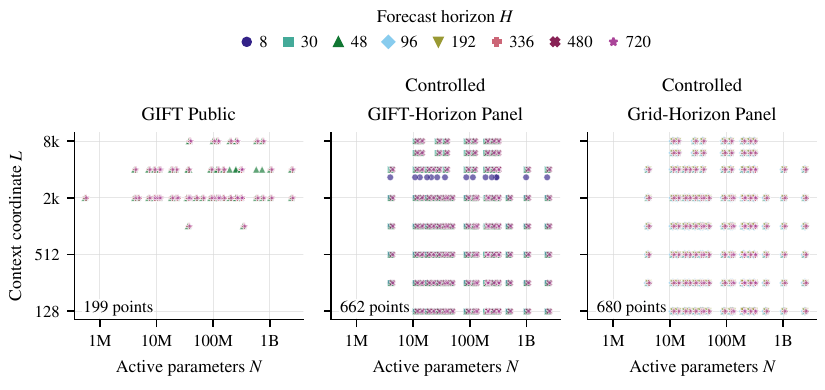}
\caption{Resource coverage across active capacity $N$, input length $L$, and scored horizon $H$ (colors and markers). Public coordinates are grid-binned; Controlled coordinates use data-supported inputs. The Controlled GIFT-Horizon Panel shows 662 protocol points and contributes 390 additional points after Grid-first selection (Table~\ref{tab:evidence}). Small horizontal offsets separate overlapping horizons.}
\label{fig:coverage}
\end{figure}

\paragraph{Long-context support.}
\label{app:long_context_support}
The Controlled input grid ends at 8192, an allocation supported by nine of the 21 registered checkpoint interfaces. Under the shared-origin protocol, 12 of the 23 dataset-frequency tasks provide at least 8192 preceding observations at every scored window, and 13 provide them at one or more windows. Shorter histories are cropped per window, so an aggregated long-context coordinate can include datasets with shorter available histories. Appendix~\ref{app:balanced_context} compares fixed task and checkpoint sets with complete recorded input support.

\subsection{Shared-origin horizon construction}
\label{app:origins}
Both controlled protocols scan allocated input lengths $L\in\{128,256,512,1024,2048,4096,6144,8192\}$ within each checkpoint's registered context limit. The Controlled GIFT-Horizon Panel uses the 61 official configurations, while the Controlled Grid-Horizon Panel scores $H\in\{48,96,192,336,720\}$ at shared origins across the same 23 dataset-frequency pairs. The additional GIFT horizons $H=8,30,480$ cover one, three, and nineteen pairs, respectively.

For a benchmark configuration with shortest series length $n_{\min}$ and scored horizon $H$, GIFT's window count $W(H)$ is
\begin{equation}
W(H)=\min\!\left(20,\max\!\left(1,\left\lceil0.1n_{\min}/H\right\rceil\right)\right).
\end{equation}
Let $n_i$ denote the length of series $i$ and $t_{i,k}$ its cutoff for window $k$. The Controlled GIFT-Horizon Panel retains the official $H$, window count, and origin spacing, giving $t_{i,k}=n_i-W(H)H+kH$ for $k=0,\ldots,W(H)-1$. The Controlled Grid-Horizon Panel instead uses the common window count $W^*=W(720)$ and
\begin{equation}
t_{i,k}=n_i-W^*720+k720,
\qquad k=0,\ldots,W^*-1.
\label{eq:shared_origins}
\end{equation}
Each horizon scores the first $H$ observations after these shared cutoffs, with Seasonal Naive evaluated at the same horizon and origins. Full-shot evaluation reuses this construction for $H=96,192,336$.

Input cropping is applied independently at each window, with historical observations and past-only covariates sharing the same time boundary. Each checkpoint retains its native forecasting path, including direct prediction, internal recursion, or masked-future generation; $H$ denotes the final scored interval.

\subsection{Response construction and display coordinates}
\label{app:response_points}
Cells are identified by checkpoint, dataset-frequency, scored horizon, and allocated input length. Before aggregation, Grid-first selection retains the Grid observation at each shared key, including its forecast origins and matched baseline, at overlapping horizons $H=48,720$. Retained dataset-cell responses are averaged equally within each checkpoint, horizon, and allocated input condition, using its data-supported long-end context. Repeated actual NHL coordinates are merged within each checkpoint; checkpoints with equal parameter counts retain separate identities. Logarithms are taken after response averaging, and final response points have equal weight in prediction-error summaries.

Continuous fitting and prediction use data-supported actual context lengths; predefined input-length levels organize discrete displays and matched local comparisons. Figure~\ref{fig:response}b groups capacity into five intervals bounded by $\{4,16,64,256,1024,4096\}$M. The capacity and horizon panels group curves by input length, while the input-length panel groups curves by horizon. Each panel uses all Controlled response points: predictions are evaluated at individual actual NHL coordinates, then averaged with the same membership and weights as the observations.

Public configurations retain the original seven-level context grid $\{128,256,512,1024,2048,4096,8192\}$ and original horizon bins. For cross-source overlap alone, Controlled coordinates are mapped onto that same Public grid; shared checkpoint--horizon-bin--context-bin keys are excluded from Public evaluation.

\subsection{Empirical relations and local targets}
\label{app:derivation}
Suppose $f$ is differentiable on a connected resource region and obeys Eqs.~\ref{eq:capacity}--\ref{eq:horizon}. With integration function $g(h,\ell)$, the capacity relation gives $f=-\beta n\ell+g(h,\ell)$. Substituting into the context relation yields $\partial g/\partial\ell=-\gamma+2q\ell$. Introducing the remaining integration function $a(h)$ gives $g=-\gamma\ell+q\ell^2+a(h)$. Finally, the horizon relation gives $a(h)=\alpha h+c$, recovering Eq.~\ref{eq:law}.

Before constructing local contrasts, points sharing a checkpoint, exact horizon, and input-length level are averaged in log response and actual context. Horizon-elasticity construction uses the corresponding mean log response. Four registered multi-capacity checkpoint subsets supply local capacity comparisons: Toto~2.0 (all five sizes), Moirai~1 (small, base, large), Chronos~2 (small, base), and Chronos-Bolt (small, base). Within each subset, an ordinary least-squares slope of $\log R$ against $\log N$ estimates capacity elasticity at each common horizon--input-length condition containing at least two checkpoints. Indexing capacity targets by $i$, slopes at an input-length level are pooled across the available subsets and horizons; their median becomes the capacity target $u_i$, paired with their mean actual context $L_i$.

For each checkpoint and horizon, consecutive observed input-length levels produce changes in log response. The actual endpoint contexts and log-response changes are averaged over available horizons separately for each checkpoint and input-length transition. Indexing context targets by $j$, this gives log-response-change targets $v_j$ at active capacity $N_j$ and endpoints $L_j^-,L_j^+$. Let $K$ and $J$ count capacity and context targets, respectively. The complete fit uses $K=8$ and $J=115$; the leave-Toto-out fit uses $K=8$ and $J=90$. The predicted capacity and context targets, $\widehat u_i$ and $\widehat v_j$, are
\begin{align}
\widehat u_i&=-\beta\log L_i,\\
\widehat v_j&=-(\gamma+\beta\log N_j)\log(L_j^+/L_j^-)
+q\big[(\log L_j^+)^2-(\log L_j^-)^2\big].
\end{align}

\subsection{Parameter estimation and prediction metrics}
\label{app:estimation}
The residual vector concatenates capacity residuals $r_i^{(N)}$ and context residuals $r_j^{(L)}$:
\begin{equation}
r_i^{(N)}=\frac{2(\widehat u_i-u_i)}{\sqrt K},
\qquad
r_j^{(L)}=\frac{\widehat v_j-v_j}{\sqrt J}.
\label{eq:residuals}
\end{equation}
The factor two scales capacity residuals before robust loss evaluation. Let $a$ index the concatenated residuals $r_a$, let $s=0.02$ be the robust-loss scale, and define the soft-$\ell_1$ loss $\rho(z)=2(\sqrt{1+z}-1)$ for a squared standardized residual $z$. Optimization uses the objective
\begin{equation}
\min_{(\gamma,\beta,q)\in\mathbb R^3}\ \frac{s^2}{2}\sum_a
\rho\!\left((r_a/s)^2\right).
\end{equation}

Within each checkpoint and input-length level in the Controlled Grid-Horizon Panel, index the sorted scored horizons by $k$ and define adjacent elasticities $a_k$:
\begin{equation}
a_k=\frac{\log R(H_{k+1})-\log R(H_k)}{\log H_{k+1}-\log H_k}.
\end{equation}
Hats denote parameter estimates. The estimate $\widehat\alpha$ is the median of these elasticities. For final combined Controlled points indexed by $p$, write $(N_p,H_p,L_p,R_p)$ for each point's resource coordinates and observed response. The intercept estimate is then
\begin{equation}
\widehat c=\operatorname{median}_p\left[
\log R_p-\widehat\alpha\log H_p+\widehat\gamma\log L_p
+\widehat\beta\log N_p\log L_p-\widehat q(\log L_p)^2\right].
\end{equation}
The leave-Toto-out procedure removes every Toto~2.0 observation from both the response and horizon-evidence panels before reconstructing the local targets and re-estimating all five coefficients. In the dataset-held-out comparisons, response aggregation and local-target construction are repeated within each training partition, with validation responses aggregated separately (Appendix~\ref{app:comparison}).

For $P$ response points, let $y_p=\log R_p$ and $\widehat y_p=\log\widehat R_p$ denote observed and predicted log responses. The absolute percentage error (APE) and its mean MAPE are
\begin{align}
\mathrm{APE}_p&=100\left|\exp(\widehat y_p-y_p)-1\right|,\\
\mathrm{MAPE}&=P^{-1}\sum_p\mathrm{APE}_p.
\end{align}
Median APE is the empirical median of the pointwise errors. The Toto capacity-curve evaluation additionally averages observed and predicted responses over horizons at each fixed capacity and context, then computes APE on the five resulting capacity means.

\section{Additional experimental results of the Unified Scaling Law}
\label{app:results}

\subsection{Reference predictions and estimation stability}
\label{app:prediction_stability}
Table~\ref{tab:prediction_references} compares the Unified law with a constant prediction, set to the exponential of the training-side median log response, and a capacity-independent form obtained by fixing $\beta=0$ and re-estimating the remaining coefficients through the same procedure. All forms use the same evaluation points and exclude every Toto checkpoint when fitting their Toto predictions.

\begin{table}[htbp]
\centering
\small
\caption{Reference prediction MAPEs (\%) on Controlled, Public, and held-out Toto response points.}
\label{tab:prediction_references}
\begin{tabular}{lrrr}
\toprule
Form & Controlled & Public & Toto \\
\midrule
Constant & 14.83 & 16.92 & 17.10 \\
Capacity-independent & 9.96 & 13.01 & 11.84 \\
Unified & 10.13 & 12.82 & 11.53 \\
\bottomrule
\end{tabular}
\end{table}

For coefficient and prediction stability, we resample the 15 base datasets with replacement 1,000 times, retaining all frequencies and observations of each draw, and repeat aggregation and every estimation step with the checkpoint panel fixed. The 95\% percentile intervals are $[0.00090,0.00322]$ for $\beta$ and $[0.0131,0.0310]$ for $q$, supporting history-dependent capacity gains and diminishing context returns.

Prediction comparisons use paired resamples for the Unified and capacity-independent forms, excluding all Toto checkpoints in each Toto fit. Evaluated on the fixed original response points, the 95\% bootstrap ranges for Unified MAPE are $[9.59,18.90]\%$ for Controlled, $[12.45,21.05]\%$ for Public, and $[10.84,20.11]\%$ for Toto. Paired MAPE-difference intervals include zero.

\subsection{Complete Toto held-out evaluation}
\begin{figure}[htbp]
\centering
\includegraphics[width=\linewidth]{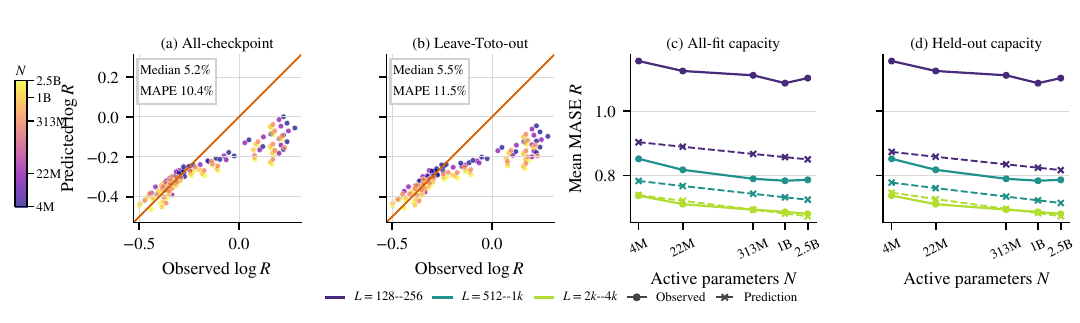}
\caption{Complete Toto~2.0 evaluation. (a,b) Observed versus predicted $\log R$ at all 150 original NHL coordinates for the all-checkpoint and leave-Toto-out fits; color encodes active parameters and the diagonal denotes exact prediction. (c,d) Capacity response after pairing adjacent input lengths and averaging the five horizons. Solid circles are observed Toto means and dashed crosses are predictions from the corresponding fit. Each context curve contains all five Toto checkpoints.}
\label{fig:toto_full}
\end{figure}

Figure~\ref{fig:toto_full} extends the main-text Toto evaluation to all 150 responses in the Controlled Grid-Horizon Panel, showing pointwise predictions and capacity profiles for the all-checkpoint and leave-Toto-out fits.

\subsection{Context-gain calculation and complete-support validation}
\label{app:resources}
\label{app:balanced_context}
For each fixed checkpoint and horizon, let $R_{\mathrm{left}}$ and $R_{\mathrm{right}}$ denote responses at shorter and longer inputs sharing the baseline denominator. The paired context gain in Figure~\ref{fig:resources} is $100(R_{\mathrm{left}}-R_{\mathrm{right}})/R_{\mathrm{left}}$; positive values indicate improvement. The fitted curves illustrate the broad saturation region, with the 100M fit varying by about 1\% across 4k--8k.

Using history and interface metadata, we select eight checkpoints and 12 tasks with complete recorded input support across all eight input lengths and five shared-origin Grid horizons, fixing task membership, forecast origins, and checkpoint membership. On this support set, mean paired gains are 3.46\% for 1024--2048 and 1.33\% for 2048--4096, becoming slightly negative for the two extensions beyond 4096. Repeating the estimation procedure gives saturation scales of approximately 3.6k--4.6k over 10M--1B parameters.

\subsection{Qualitative restrictions of reduced laws}
\label{app:form_geometry}

Figure~\ref{fig:reduced_law_properties} applies the restrictions in Section~\ref{sec:form_selection} to the full-panel law without refitting. Panel (a) uses adjacent-capacity checkpoints from the registered subsets at common horizons and input-length level $L=1024$. Panel (b) subtracts the observed $L=1024$ response separately within each checkpoint--horizon slice for the three checkpoints between 75M and 150M parameters. Panel (c) converts the local capacity slopes pooled across heterogeneous checkpoint subsets to tenfold log-response changes. Panel (d) divides adjacent-context log-response changes by their actual log-input change $\Delta\log L$ for checkpoints with $N\leq50$M, whose median log capacity defines the representative 22M curve; the finite difference equals the quadratic law's derivative at the geometric midpoint.

\subsection{Flexible alternatives and common estimation}
\label{app:comparison}

We compare the Unified law with four alternatives that can express changing context returns and capacity--history coupling: the Unified law with an independent-capacity term (U+N), conditional quadratic (CQ), rational saturation (RS), and capacity-gated saturation (CGS), with log responses $f_{\mathrm{U+N}}$, $f_{\mathrm{CQ}}$, $f_{\mathrm{RS}}$, and $f_{\mathrm{CGS}}$, respectively. Within these forms, $s>0$ is a context scale in CQ and CGS; $u$ and $v$ control capacity and horizon dependence, respectively. In RS, $L_s>0$ is the context saturation scale; in CGS, $N_0>0$ is the capacity-gate scale. The forms are
\begin{align}
f_{\mathrm{U+N}}={}&c+\alpha h+\lambda n-\gamma\ell-\beta n\ell+q\ell^2,\\
f_{\mathrm{CQ}}={}&c+\lambda n+\alpha h+q(\ell-\log s-un-vh)^2,\\
f_{\mathrm{RS}}={}&c+\alpha h-(\gamma+\beta n)\frac{L}{L+L_s},\\
f_{\mathrm{CGS}}={}&c+\alpha h-\gamma\frac{N}{N+N_0}
\left[1-\exp\!\left(-\frac{L}{sH^v}\right)\right].
\end{align}
The first is the six-parameter nested extension. CQ uses a capacity- and horizon-dependent quadratic center and has seven parameters. RS has five parameters and an explicit context scale. CGS has six parameters and combines a Hill capacity gate with a horizon-dependent saturation scale. Parameter counts include $c$ and $\alpha$.

The comparison uses the 23-task Controlled Grid-Horizon Panel and its 16 development checkpoints; the complete Toto~2.0 checkpoint series is excluded. The 15 base datasets form three fixed folds, keeping every frequency and checkpoint--horizon--context observation of a base dataset in the same fold. Training and validation responses are aggregated separately. Each candidate fits its implied local relations using the same targets, weighting, and robust objective; the intercept is calibrated last. Optimization and initialization use training data only.

Mean validation MAPEs across the three folds are 11.77\% for Unified, 11.78\% for its independent-capacity extension, 11.87\% for CQ, 11.90\% for RS, and 13.10\% for CGS. Fold assignments and per-fold results are provided in the released analysis.

\subsection{Horizon-interaction extensions}
\label{app:horizon_interactions}
\label{app:interaction_evidence}

Let $f_{\mathrm U}$ denote the Unified log-response law in Eq.~\ref{eq:law}, and let $\eta$ and $\kappa$ be capacity--horizon and context--horizon interaction coefficients. We assess the common-horizon-shift specification with the nested family
\begin{equation}
f=f_{\mathrm U}+\eta nh+\kappa\ell h.
\label{eq:horizon_interaction_extension}
\end{equation}
We allow $\eta$ alone, $\kappa$ alone, or both. These extensions have six, six, and seven parameters. The $nh$ term permits horizon-dependent capacity gains, while $\ell h$ permits a horizon-dependent context saturation scale.

All forms use the same dataset folds and jointly fit matched capacity, context, and horizon contrasts. Validation errors remain comparable across these forms.

\section{Unified Theory of Time Series Learning: theoretical analysis}
\label{app:learning_theory}

\subsection{Gaussian learning model}
\label{app:learning_regression}

We analyze a fixed query $B$ and scalar forecast target $Y$. Conditional on this query, let
\begin{equation}
Y=\sum_{j=1}^{p}\psi_j(B)\omega_j+\varepsilon,
\qquad
\omega_j\sim\mathcal N(0,\tau_j^2),
\qquad
\varepsilon\sim\mathcal N(0,\sigma^2).
\label{eq:learning_gaussian_target}
\end{equation}
The query features $\psi_j(B)$ are fixed, and the regression coefficients and innovation are independent. Let $z_{L,j}$ denote the sufficient statistic supplied by historical observations for coefficient $j$, with independent standard Gaussian noises $\xi_j$ also independent of the coefficients and innovation:
\begin{equation}
z_{L,j}=\omega_j+\frac{\xi_j}{\sqrt{L\nu_j}},
\qquad
\xi_j\sim\mathcal N(0,1),\quad \nu_j>0,
\label{eq:learning_evidence}
\end{equation}
Thus $L\nu_j$ is the precision of the historical evidence about coefficient $j$. Gaussian conditioning gives the posterior mean $\widehat\omega_j(L)$ and variance $v_j(L)$ \citep{rasmussen2006gaussian}:
\begin{equation}
\widehat\omega_j(L)=\frac{L\nu_j\tau_j^2}{1+L\nu_j\tau_j^2}z_{L,j},
\qquad
v_j(L)=\frac{\tau_j^2}{1+L\nu_j\tau_j^2}.
\label{eq:learning_posterior}
\end{equation}

Consider nested predictors that use the posterior means of the first $r$ coefficients and the prior means of the remainder. Here $r$ counts the available regression coefficients. The coefficients $\omega_j$ describe the dynamic rule; the fitted or pretrained neural parameters determine how an actual model estimates and applies such a rule. Under the joint distribution of coefficients, evidence, and innovation at fixed $B$, the prediction error is a centered Gaussian with variance
\begin{equation}
V_r(L)=\sigma^2+\sum_{j\le r}\psi_j(B)^2v_j(L)
                 +\sum_{j>r}\psi_j(B)^2\tau_j^2.
\label{eq:learning_variance_appendix}
\end{equation}
The posterior mean is also the posterior median, so the minimum expected absolute error among predictors using these $r$ statistics is
\begin{equation}
\mathcal E_r(L)=\sqrt{\frac{2}{\pi}}\sqrt{V_r(L)}.
\label{eq:learning_absolute_error}
\end{equation}
The estimate can be retained after training or computed from current history, as in in-context regression of autoregressive rules \citep{sander2024incontext}.

\subsection{Historical evidence and available rule coefficients}
\label{app:learning_returns}

Write $a_j=\psi_j(B)^2\tau_j^2$ and $b_j=\nu_j\tau_j^2$. Coefficients with $a_j=0$ do not affect the fixed query and can be omitted. Define the remaining variance $A_r=\sigma^2+\sum_{j>r}a_j$. Then
\begin{equation}
V_r(L)=A_r+\sum_{j\le r}\frac{a_j}{1+b_jL}.
\label{eq:learning_remaining_variance}
\end{equation}
We assume $A_r>0$ for the predictors under comparison. This remaining variance includes innovation and the contribution of coefficients outside the predictor's available set. Denote the variance reduction from adding coefficient $j=r+1$ by $g_j(L)$:
\begin{equation}
g_j(L)=V_r(L)-V_{r+1}(L)=\frac{a_jb_jL}{1+b_jL}.
\label{eq:learning_coefficient_gain}
\end{equation}
Its absolute-error improvement is
\begin{equation}
\mathcal E_r(L)-\mathcal E_{r+1}(L)
=\sqrt{\frac{2}{\pi}}\,
\frac{g_j(L)}{\sqrt{V_r(L)}+\sqrt{V_{r+1}(L)}}.
\label{eq:learning_capacity_gain}
\end{equation}
The numerator increases with $L$, while both variances in the denominator are nonincreasing. The improvement therefore increases with historical evidence whenever the added coefficient affects the query.

For fixed $r$ and finite $p$, put $B_r=\sum_{j\le r}a_j/b_j$. Writing $O(L^{-2})$ for a remainder bounded in magnitude by a constant times $L^{-2}$ at large $L$, expanding Eq.~\ref{eq:learning_remaining_variance} gives
\begin{equation}
V_r(L)=A_r+\frac{B_r}{L}+O(L^{-2}),
\qquad
\mathcal E_r(L)=\sqrt{\frac{2A_r}{\pi}}
\left(1+\frac{B_r}{2A_rL}+O(L^{-2})\right).
\label{eq:learning_long_history}
\end{equation}
The finite-history estimation error decreases while the remaining variance dominates. In particular, the magnitude of the log-context derivative of error approaches zero.

\subsection{Retained rule information and immediate evidence}
\label{app:learning_substitution}

For one coefficient, suppose training leaves a Gaussian posterior with variance $v>0$, and immediate historical evidence contributes an additional Gaussian likelihood with conditional precision $I\ge0$. Gaussian conditioning gives the updated posterior variance $v_{\mathrm{post}}$:
\begin{equation}
v_{\mathrm{post}}=\frac{v}{1+Iv}.
\label{eq:learning_retained_posterior}
\end{equation}
Here $I$ measures information additional to the retained estimate, accounting for overlap between the training record and current history by conditioning on retained evidence. For fixed query weight $k>0$ and remaining variance $A\ge0$, the absolute-error gain $G(v;I)$ from that immediate evidence is
\begin{equation}
G(v;I)=\sqrt{\frac{2}{\pi}}\left[
\sqrt{A+kv}-\sqrt{A+\frac{kv}{1+Iv}}\right].
\label{eq:learning_substitution_gain}
\end{equation}

With $I$, $k$, and $A$ fixed, this gain is nondecreasing in $v$. Indeed,
\begin{equation}
\frac{\partial G}{\partial v}
=\frac{k}{\sqrt{2\pi}}\left[
\frac{1}{\sqrt{A+kv}}
-\frac{1}{(1+Iv)^2\sqrt{A+kv/(1+Iv)}}\right]\ge0,
\label{eq:learning_substitution_proof}
\end{equation}
because $(1+Iv)^4[A+kv/(1+Iv)]\ge A+kv$, with strict inequality for $I>0$. More accurate retention of the same coefficient therefore reduces the gain from additional immediate evidence. Comparing two positive precisions follows by first incorporating the smaller precision into $v$ and applying the same argument to the additional precision.

\section{Unified Theory of Time Series Learning: experimental evidence}
\label{app:mechanism}

\subsection{History budgets, immediate input, and adaptation}
\label{app:mechanism_history}
\label{app:mechanism_crossed}

The matched-history comparison uses DLinear and PatchTST adapted from Time-Series-Library \citep{wu2023timesnet}. A shared univariate model is trained on each task's scalar histories, using the available length-$L$ record before the earliest scoring origin, 96-step inputs, and $H$-step targets. Adam uses learning rate $10^{-4}$, batch size 32, and three initialization seeds. Chronological validation selects an epoch count, followed by refitting on the complete pre-origin record; weights remain fixed during scoring. The comparison contains 1,656 full-shot cells. Each plotted response averages the 23 tasks and then the seeds equally; seed variability is the sample standard deviation relative to the seed mean at each horizon--history condition. TSFM curves summarize the same nine checkpoints covering the complete grid, each represented by its 23-task arithmetic mean. Remaining architecture, normalization, and optimization settings are specified in the released configuration.

The crossed experiment varies training history $T\in\{4096,8192\}$ and immediate input $C\in\{96,512\}$ at $H=96$ on the same 23 tasks. DLinear, PatchTST, a compact PatchTST, and pretrained TimesFM 2.5 each use three seeds, giving \MechanismCrossedFitCount{} fits. Histories end before the earliest scoring origin. Training starts and validation targets are matched across input lengths by cropping the corresponding 512-step windows for shorter inputs; evaluation origins and targets are identical across arms. Each input configuration is trained separately under a common update budget, with validation selecting among the initial and subsequent predictors without refitting. MASE uses the same GIFT scoring and matched Seasonal Naive reference as the main evaluation. Tasks are averaged equally, with seeds and the balanced history--input cells averaged within task.

\begin{table}[htbp]
\caption{Validation-selected MASE in the crossed experiment, averaged over 23 tasks and three seeds. $T$ is training history and $C$ is immediate input.}
\label{tab:mechanism_crossed}
\centering
\small
\begin{tabular}{lrrr}
\toprule
Model & Input $C$ & History $T=4096$ & History $T=8192$ \\
\midrule
DLinear & 96 & 1.04264 & 1.03958 \\
DLinear & 512 & 0.88869 & 0.88328 \\
PatchTST & 96 & 0.91400 & 0.91361 \\
PatchTST & 512 & 0.86638 & 0.84896 \\
PatchTST-compact & 96 & 0.99099 & 0.98986 \\
PatchTST-compact & 512 & 0.90357 & 0.89159 \\
TimesFM & 96 & 0.83326 & 0.81896 \\
TimesFM & 512 & 0.74498 & 0.75370 \\
\bottomrule
\end{tabular}

\end{table}

Adapted TimesFM improves MASE on 20 tasks, is unchanged on two, and increases it on one. Averaging training histories and seeds within task, frozen TimesFM has MASE 1.0431 and 0.8035 at inputs 96 and 512; the adapted configurations have MASE 0.8261 and 0.7493. The short--long input gap decreases from 0.2396 to 0.0768. Its reduction of 0.1628 has a 95\% bootstrap interval of $[0.0170,0.3248]$, obtained from 10,000 resamples of the 15 base datasets while keeping their frequency tasks together.

\subsection{Controlled processes, evidence splits, and scoring}
\label{app:mechanism_design}

Writing $\operatorname{sign}$ for the sign function, the controlled processes with rule sign $s\in\{-1,+1\}$, magnitude $a\in\{0.15,0.225,0.3\}$, and independent standard Gaussian innovations $\epsilon_t$ are
\begin{align}
\text{AR(1):}\quad x_t&=sa\,x_{t-1}+\sqrt{1-a^2}\epsilon_t,\\
\text{Lag-8:}\quad x_t&=sa\,x_{t-8}+\sqrt{1-a^2}\epsilon_t,\\
\text{Threshold:}\quad x_t&=sa\,\operatorname{sign}(x_{t-1})+\sqrt{1-a^2}\epsilon_t.
\end{align}
AR(1) and lag-8 have standard Gaussian stationary marginals; lag-8 starts with eight independent standard Gaussian observations. The threshold process starts from its stationary equal mixture of $\mathcal N(-a,1-a^2)$ and $\mathcal N(a,1-a^2)$. Every process has mean zero and variance one, and the two signs have the same marginal distribution within each process and magnitude.

Independent 96-step queries are drawn from the equal mixture of the two stationary rules and retained when their absolute rule log-likelihood ratio is at most $\log3$, selecting queries that leave uncertainty about the rule. Each query has an antithetic partner obtained by negating the entire series. Frozen-context and parameter experiments use 128 independent development queries per process at $a=0.3$ and 1,024 independent test queries per process and magnitude. Historical prefixes are sampled from the exact reverse conditional process given the query boundary; frozen-context comparisons use nested lengths and four prefix realizations. Activation experiments use separate query streams, with 128 development and 1,024 confirmation queries per process and magnitude, and two donor repetitions. Development and confirmation donors are independent; the primary TimesFM setting and candidate grids are fixed before confirmation.

Let $b_j$, $j=1,\ldots,96$, denote the query observations. The primary target is the one-step conditional mean: $sa\,b_{96}$ for AR(1), $sa\,b_{89}$ for lag-8, and $sa\,\operatorname{sign}(b_{96})$ for threshold. MASE follows Eq.~\ref{eq:relative}, with both model and naive forecasts scored against this target. The raw absolute-error scale is the mean lag-one absolute difference within the common 96-step query; the naive forecast is $b_{96}$. Raw scores are pooled across paired rule signs before division by the naive score, then averaged across conditions. Paired comparisons retain antithetic queries in the same statistical unit and average prefix-level errors when prefixes are repeated. The 95\% intervals use 2,000 bootstrap resamples of independent queries, preserving their paired predictions and histories. Training seeds are reported separately; query intervals condition on the fitted models and donor states. The decomposition in Eq.~\ref{eq:mechanism_decomposition} uses the same predictions.

\subsection{Historical evidence and rule identification}
\label{app:mechanism_context}
\label{app:mechanism_oracle}

Chronos 2 and TimesFM 2.5 are evaluated at context lengths 96, 128, 256, 512, 1024, and 2048. Each emits its native median forecast from the first output block, with no rollout or state shared across examples. We compare native normalization statistics with fixed population location zero and scale one; Chronos retains its nonlinear scaling transform. Interventions use fixed statistics throughout. Using subscripts $+$ and $-$ for rule signs $s=+1$ and $s=-1$, respectively, denote predictions by $f_+(B),f_-(B)$ and conditional means by $\mu_+(B),\mu_-(B)$. Summing over evaluated queries $B$, define the rule projection $P$:
\begin{equation}
P=\frac{\sum_B [f_+(B)-f_-(B)][\mu_+(B)-\mu_-(B)]}
        {\sum_B [\mu_+(B)-\mu_-(B)]^2}.
\label{eq:mechanism_projection}
\end{equation}
It measures response direction and amplitude, with one corresponding to the oracle contrast amplitude. Table~\ref{tab:mechanism_context} shows prediction gains and differences between models' rule responses. Across both normalization conditions, all \MechanismFrozenConditionCount{} model--process--magnitude--normalization combinations have positive intervals for the MASE gain from 96 to 2048.

\begin{table}[htbp]
\caption{Conditional-mean MASE with fixed population statistics at $a=0.3$. Rule projection measures the predicted contrast along the known conditional-mean contrast; one corresponds to the oracle amplitude.}
\label{tab:mechanism_context}
\centering
\small
\begin{tabular}{llrrr}
\toprule
Process & Frozen model & $L=96$ & $L=2048$ & Rule projection \\
\midrule
AR(1) & TimesFM 2.5 & 0.32424 & 0.12650 & 0.812 \\
AR(1) & Chronos 2 & 0.33295 & 0.12958 & 0.666 \\
Lag-8 & TimesFM 2.5 & 0.33272 & 0.16004 & 0.739 \\
Lag-8 & Chronos 2 & 0.34423 & 0.26440 & 0.133 \\
Threshold & TimesFM 2.5 & 0.37300 & 0.22949 & 0.454 \\
Threshold & Chronos 2 & 0.37281 & 0.25215 & 0.369 \\
\bottomrule
\end{tabular}

\end{table}

An oracle knows the process family and coefficient magnitude and estimates the rule sign from the historical prefix, using the same queries, conditional histories, and scoring as the frozen models. Write $D_L$ for the history $D_{L,t}$ containing query $B$, with the forecast-origin index suppressed, and let $p_+$ and $p_-$ denote sequence densities under the positive and negative rules. Because both signs are equally weighted for each fixed query $B$, the log posterior odds $\Lambda(D_L\mid B)$ are
\begin{equation}
\Lambda(D_L\mid B)=\log\frac{p_+(D_L)}{p_-(D_L)}
                         -\log\frac{p_+(B)}{p_-(B)}.
\label{eq:mechanism_oracle_odds}
\end{equation}
Selecting the more probable sign gives a posterior-median prediction and minimizes expected absolute error; ties select the positive sign. At the smallest magnitude, $a=0.15$, its MASE is below 0.001 for all three processes at $L=512$. All processes and magnitudes have zero sign errors at lengths 1,024 and 2,048 in the registered finite sample. The released results contain all 54 process--magnitude--length conditions.

\subsection{Parameter retention and transfer}
\label{app:mechanism_weights}

DLinear and a compact PatchTST learn the controlled processes with fixed population normalization, 96 inputs, three outputs, and mean squared error (MSE) loss. Histories have lengths 1024, 4096, or 8192, with three training seeds and opposite-rule pairs, giving \MechanismFitCount{} individual fits. Histories are nested, and opposite rules share initialization and sampling randomness. An 80\% chronological training split supplies windows; the remaining targets select the predictor without refitting. Architecture and optimizer settings are specified in the released configuration.

Trained models predict independent queries after the training record has left the forecast computation. We exchange parameters between opposite-rule models and calibrate their batch-normalization (BatchNorm) statistics, where present, on the same independent unlabelled mixture of historical windows for both directions, without future targets or parameter updates. Error to the donor rule decreases in all \MechanismParameterConditionCount{} process--architecture--length--seed conditions. Both architectures' mean prediction errors decrease with training-history length (Table~\ref{tab:mechanism_weights}).

\begin{table}[htbp]
\caption{Conditional-mean MASE of trained predictors, averaged over three seeds. Transfer projection measures the donor-rule response after parameter exchange and shared-history normalization calibration at $L=8192$.}
\label{tab:mechanism_weights}
\centering
\small
\setlength{\tabcolsep}{4pt}
\begin{tabular}{llrrrr}
\toprule
Process & Model & $L=1024$ & $L=4096$ & $L=8192$ & Transfer projection \\
\midrule
AR(1) & DLinear & 0.27871 & 0.16235 & 0.12634 & 0.986 \\
AR(1) & PatchTST & 0.32585 & 0.14790 & 0.09567 & 1.028 \\
Lag-8 & DLinear & 0.32463 & 0.17748 & 0.12204 & 0.985 \\
Lag-8 & PatchTST & 0.34635 & 0.14133 & 0.10034 & 0.946 \\
Threshold & DLinear & 0.35156 & 0.25287 & 0.22841 & 0.584 \\
Threshold & PatchTST & 0.37115 & 0.20446 & 0.18716 & 0.755 \\
\bottomrule
\end{tabular}

\end{table}

\subsection{Activation transfer, recovery, and controls}
\label{app:mechanism_transfer}
\label{app:mechanism_specificity}
\label{app:mechanism_decomposition}

The primary TimesFM 2.5 intervention uses activation patching \citep{zhang2024patching} at the sixth Transformer block's residual output, over the final three query tokens, with input length 512 and unit strength. These settings were fixed on development evidence before independent confirmation. Each recipient uses four independent 512-step donor histories from the same process and rule, with two donor repetitions. Chronos 2 uses the same donor construction and history lengths, patching the six tokens corresponding to the final 96 observations while leaving its register and forecast tokens unchanged. Its layer and strength are selected on development evidence as specified below.

For donor index $j\in\{1,\ldots,4\}$, let $D_j$ denote donor history and $B_j$ its final 96 observations. Let $h$ map each input to the selected donor-token states. We construct the mean state difference $\Delta h$:
\begin{equation}
\Delta h=\frac14\sum_{j=1}^4\bigl[h(D_j)-h(B_j)\bigr].
\label{eq:mechanism_patch}
\end{equation}
Only observed donor histories enter this construction; known synthetic rules determine matched and opposite donor assignment and scoring targets. With intervention strength $\lambda$, transfer replaces a new query's short-input state by $h(B)+\lambda\Delta h$. Recovery first replaces the selected state in its natural 512-step execution by $h(B)$, defining erasure, and then adds $\lambda\Delta h$. Other long-context states, including prefixes accessible to downstream layers, remain unchanged.

Controls match donor count, strength, and location. Opposite-rule donors hold the recipient query and long history fixed and pair donor carriers and sampling randomness across signs. Random changes match the norm of $\Delta h$; carrier-only changes compare independent short donor queries; shuffled-history changes permute each donor's earlier observations while preserving its final 96 values. Natural predictions and norm-matched random erasure are also retained.

TimesFM is evaluated at blocks 4, 6, and 8 with strengths $0.5,1,1.5$, retaining all nine settings. Chronos development evaluates blocks 3, 6, and 9 at the same strengths. Equal averaging of development conditional-mean MASE across processes, magnitudes, repetitions, signs, and both objectives selects block 6 and strength 1.5, which are then fixed for independent confirmation. Each setting contains 18 process--magnitude--donor-repetition conditions sharing the registered query sets.

\begin{table}[htbp]
\caption{Conditional-mean MASE for the primary TimesFM setting and independently confirmed Chronos setting. Baseline is query-only prediction for transfer and erased prediction for recovery. Matched and opposite refer to the donor rule.}
\label{tab:mechanism_specificity}
\centering
\small
\setlength{\tabcolsep}{3pt}
\begin{tabular}{llrrrrr}
\toprule
Model & Objective & Baseline & Matched & Opposite & Random & Shuffled \\
\midrule
TimesFM 2.5 & Transfer & 0.27810 & 0.25720 & 0.30316 & 0.31468 & 0.26804 \\
TimesFM 2.5 & Recovery & 0.25821 & 0.23726 & 0.29228 & 0.29514 & 0.25137 \\
Chronos 2 & Transfer & 0.28774 & 0.28776 & 0.29987 & 0.28710 & 0.29144 \\
Chronos 2 & Recovery & 0.21064 & 0.19798 & 0.20436 & 0.20984 & 0.20454 \\
\bottomrule
\end{tabular}

\end{table}

Matching donors outperform opposite-rule donors in all 324 TimesFM comparisons across nine settings, two objectives, and 18 conditions, each with a positive paired-query interval. Their prediction changes also align with the oracle rule contrast in every condition. Mean recovery improves at all nine settings and mean direct transfer at seven. At the primary setting, MASE is lower than with shuffled-history donors in 14 of 18 conditions for each objective. The recovery percentage in Section~\ref{sec:mechanism} is mean recovered improvement divided by mean erasure loss at that setting.

Chronos recovery improves all 18 conditions, each with a positive paired interval. Matching donors outperform opposite-rule donors in all 18 conditions for each objective, with positive intervals in 15. Its direct-transfer MASE is nearly unchanged, from \MechanismChronosTransferBefore{} to \MechanismChronosTransferAfter{} (Table~\ref{tab:mechanism_specificity}).

To separate rule responses from common prediction changes, let the two conditional means at horizon 1 be $\pm q(B)$ and predictions under the corresponding history or donor rules be $f_+(B),f_-(B)$. Suppressing the fixed query argument $B$, define the common prediction component $m=(f_++f_-)/2$ and the rule-contrast component $d=(f_+-f_-)/2$. The paired-rule squared error decomposes exactly as
\begin{equation}
\frac{(f_+-q)^2+(f_-+q)^2}{2}=m^2+(d-q)^2.
\label{eq:mechanism_decomposition}
\end{equation}
The first term measures a common prediction offset and the second agreement with the rule-dependent conditional mean. Gains are relative to query-only prediction for transfer and erasure for recovery. TimesFM's rule-dependent error improves in every condition for both objectives at the primary setting, including against shuffled-history donors, with positive paired intervals. Chronos improves the mean rule component in each process for both objectives. In direct transfer, its common-component cost exceeds this gain in MSE; recovery lowers overall error.

\end{document}